\documentclass[runningheads]{llncs}
\usepackage{graphicx}
\usepackage{orcidlink}
\usepackage{cite}
\usepackage{booktabs}
\usepackage[misc]{ifsym}
\usepackage{amsmath,amssymb}
\usepackage{array}
\usepackage{multirow}
\usepackage{tabularx}

\begin{document}
\title{Generating Reading Comprehension Exercises with Large Language Models for Educational Applications}
\titlerunning{Generating Reading Comprehension Exercises with LLMs}
%
\author{
Xingyu Huang\inst{1}\orcidlink{0009-0004-7005-2997}\and
Fei Jiang\inst{2}\orcidlink{0009-0001-9495-9903}\and
Jianli Xiao\inst{1}\textsuperscript{(\Letter)}\orcidlink{0000-0002-7363-0623}
}
\authorrunning{X. Huang et al.}
%
\institute{
School of Optical-Electrical and Computer Engineering, University of Shanghai for Science and Technology, Shanghai, China \\
\email{233350741@st.usst.edu.cn}, \email{audyxiao@sjtu.edu.cn}
\and
Chongqing Academy of Science and Technology, Chongqing, China\\
\email{jiang.feiabc@163.com}}
\maketitle              
\begin{abstract}
With the rapid development of large language models (LLMs), the applications of LLMs have grown substantially. In the education domain, LLMs demonstrate significant potential, particularly in automatic text generation, which enables the creation of intelligent and adaptive learning content. This paper proposes a new LLMs framework, which is named as Reading Comprehension Exercise Generation (RCEG). It can generate high-quality and personalized English reading comprehension exercises automatically. Firstly, RCEG uses fine-tuned LLMs to generate content candidates. Then, it uses a discriminator to select the best candidate. Finally, the quality of the generated content has been improved greatly. To evaluate the performance of RCEG, a dedicated dataset for English reading comprehension is constructed to perform the experiments, and comprehensive evaluation metrics are used to analyze the experimental results. These metrics include content diversity, factual accuracy, linguistic toxicity, and pedagogical alignment. Experimental results show that RCEG significantly improves the relevance and cognitive appropriateness of the generated exercises.

\end{abstract}
\section{Introduction}
The emergence of large language models (LLMs) has opened new avenues for educational technology, particularly in English language learning~\cite{hao2024transforming,voultsiou2025systematic,kasneci2023chatgpt,wang2024large}. These models excel in natural language generation and semantic understanding. They have been widely applied in dialogue systems, automated writing, and machine translation. In recent years, researchers and educators have begun to explore the role of LLMs in English education. Most studies have focused on instructional support and personalized learning. However, existing work primarily targets teaching assistance or writing correction. Relatively little attention has been given to content generation—especially the automatic creation of test items.

Previous research has demonstrated that LLMs are highly effective in text generation. They can consistently produce coherent and well-structured passages, even in few-shot or zero-shot settings. Models like ChatGPT have already been applied to automatic question generation, showing promising early results~\cite{attali2018automatic,attali2022interactive,bulut2022automatic}. However, compared to applications such as instructional feedback and writing correction, the use of LLMs in generating assessment content—particularly reading comprehension exercises—has received relatively little attention. To address this gap, the present study investigates the feasibility of using LLMs to generate high-quality and scalable educational resources.

\begin{figure}
\centering
\includegraphics[width=\textwidth, trim=4cm 4cm 4cm 4cm, clip]{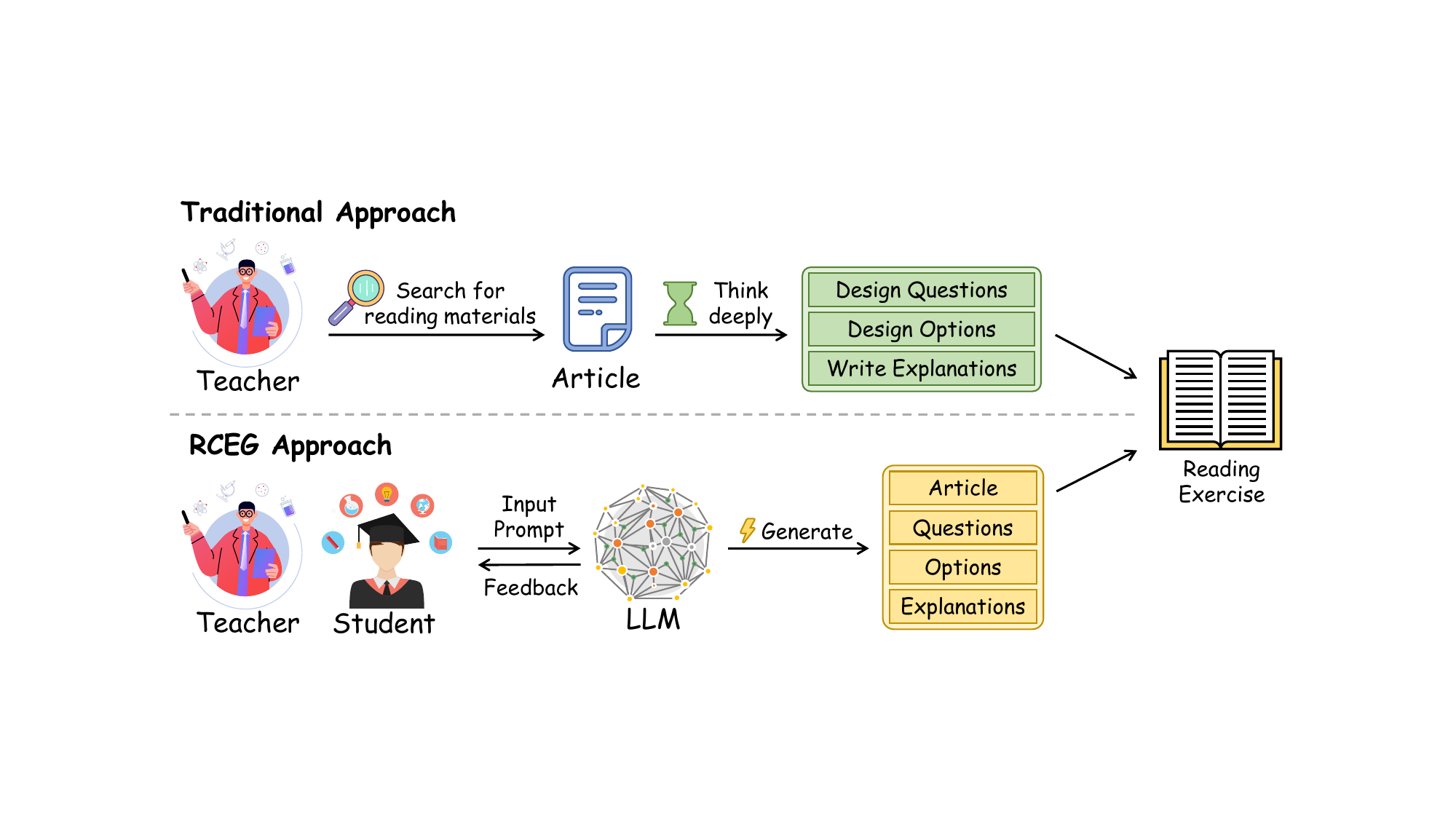}
\caption{Comparison between Traditional Reading Comprehension Exercise Design and the RCEG Framework.}
\label{fig1}
\end{figure}

To address the aforementioned challenges, we propose an automated framework for English reading instruction, termed Reading Comprehension Exercise Generation (RCEG), as illustrated in Fig.~\ref{fig1}. Built on LLMs, RCEG first applies supervised fine-tuning (SFT), followed by reinforcement learning using Proximal Policy Optimization (PPO), to train a task-specific model. To further improve content quality, the framework incorporates a post-processing module based on Dynamic Attribute Graphs-based controlled text generation (DATG)~\cite{liang2024controlled}, which generates multiple candidate outputs. These candidates are then evaluated and filtered using a discriminator from the Generative Discriminator (GeDi)\cite{krause2021gedi}, enhancing both accuracy and readability. The system is evaluated using widely adopted natural language processing (NLP) metrics across several dimensions. Experimental results show that RCEG consistently outperforms existing baselines in terms of coherence, factual accuracy, and safety. It provides educators with high-quality question resources and delivers personalized reading materials to learners. These results highlight the framework’s strong potential for advancing intelligent education systems. The key contributions of our study are summarized as follows:
\begin{itemize}
    \item We construct three task-specific datasets to support the complete instruction alignment training pipeline for educational content generation. These datasets are designed for SFT, reward model training, and reinforcement learning with PPO.
    \item We propose the Reading Comprehension Exercise Generation (RCEG) framework, which integrates a fine-tuned language model with a post-processing module. This module combines DATG for controlled generation with GeDi-based discriminator reranking. In addition, we develop a Gradio-based interactive interface to support practical use by both students and teachers.
    \item We conduct a comprehensive evaluation across multiple dimensions, including coherence, relevance, and safety. Both automatic metrics and human preference ratings are used. The results show that our framework significantly outperforms existing baselines in generating high-quality, learner-adapted reading comprehension exercises.
\end{itemize}

\section{Related Work}

\subsection{Traditional Text Generation}
Before the emergence of large language models (LLMs), a substantial body of research had already addressed the core challenges of text generation. This work laid the foundation for subsequent advances in natural language processing (NLP).

Early text generation methods relied on handcrafted grammar rules or templates, leading to rigid and domain-specific outputs with limited semantic flexibility. The introduction of statistical models marked a significant shift. N-gram models captured word sequence probabilities, while Hidden Markov Models (HMMs) incorporated state transitions to enable sequential generation~\cite{eddy1996hidden}. With the advancement of feature engineering, models such as the Maximum Entropy Model (MEM) and Conditional Random Fields (CRFs) began to integrate linguistic features—such as part-of-speech tags and syntactic structures—to improve generation accuracy~\cite{berger1996maximum,sutton2012introduction}. However, these approaches relied heavily on manual features and struggled to model long-range dependencies, which ultimately led to their replacement by deep learning methods more capable of handling complex text generation tasks.

The advent of deep learning marked a pivotal shift in text generation by enabling end-to-end modeling without relying on handcrafted features. Recurrent Neural Networks (RNNs), along with their enhanced variants such as Long Short-Term Memory (LSTM) and Gated Recurrent Units (GRUs), captured sequential dependencies through recurrent hidden states. These models achieved promising results in tasks such as machine translation and poetry generation~\cite{iqbal2022survey,li2024pre}. Building on this progress, Sutskever et al.\cite{sutskever2014sequence} introduced the encoder-decoder architecture, which framed text generation as a sequence-to-sequence mapping, improving semantic representation. Later, Vaswani et al.\cite{vaswani2017attention} proposed the Transformer architecture, which replaced recurrence with self-attention to capture global contextual dependencies. This design enabled efficient parallel computation and improved long-range sequence modeling, laying the foundation for the development of LLMs.

\subsection{Text Generation with Large Language Models}
In recent years, the rapid advancement of deep learning in NLP has led to the emergence of LLMs as the dominant paradigm for text generation. Pretrained models based on the Transformer architecture—such as BERT, GPT, T5, and LLaMA—are trained on massive unlabeled corpora using unsupervised learning. These models exhibit strong capabilities in both language understanding and generation. They have been widely applied to various tasks, including dialogue systems, text summarization, and code generation~\cite{zhao2023survey}.

In the context of educational assessment, researchers have increasingly explored the use of LLMs for generating instructional materials, particularly in automatic exercise generation. Chan et al.~\cite{chan2022agree} proposed a BERT-based system for generating distractors in grammar multiple-choice questions, achieving 85 percent expert-rated quality across 4,500 items. Zu et al.~\cite{zu2023automated} fine-tuned GPT-2 on a vocabulary test corpus and found that the generated distractors more closely resembled those written by human experts compared to rule-based methods. More recently, Xiao et al.~\cite{xiao2023evaluating} developed a ChatGPT-based system for generating reading comprehension exercises for Chinese middle school students. They conducted both human and automated evaluations. The results showed that the generated materials were not only suitable for classroom use, but in some cases even surpassed the quality of human-written textbook exercises.

\section{Methodology}
\begin{figure}
\centering
\includegraphics[width=\textwidth, trim=4cm 3cm 4cm 3cm, clip]{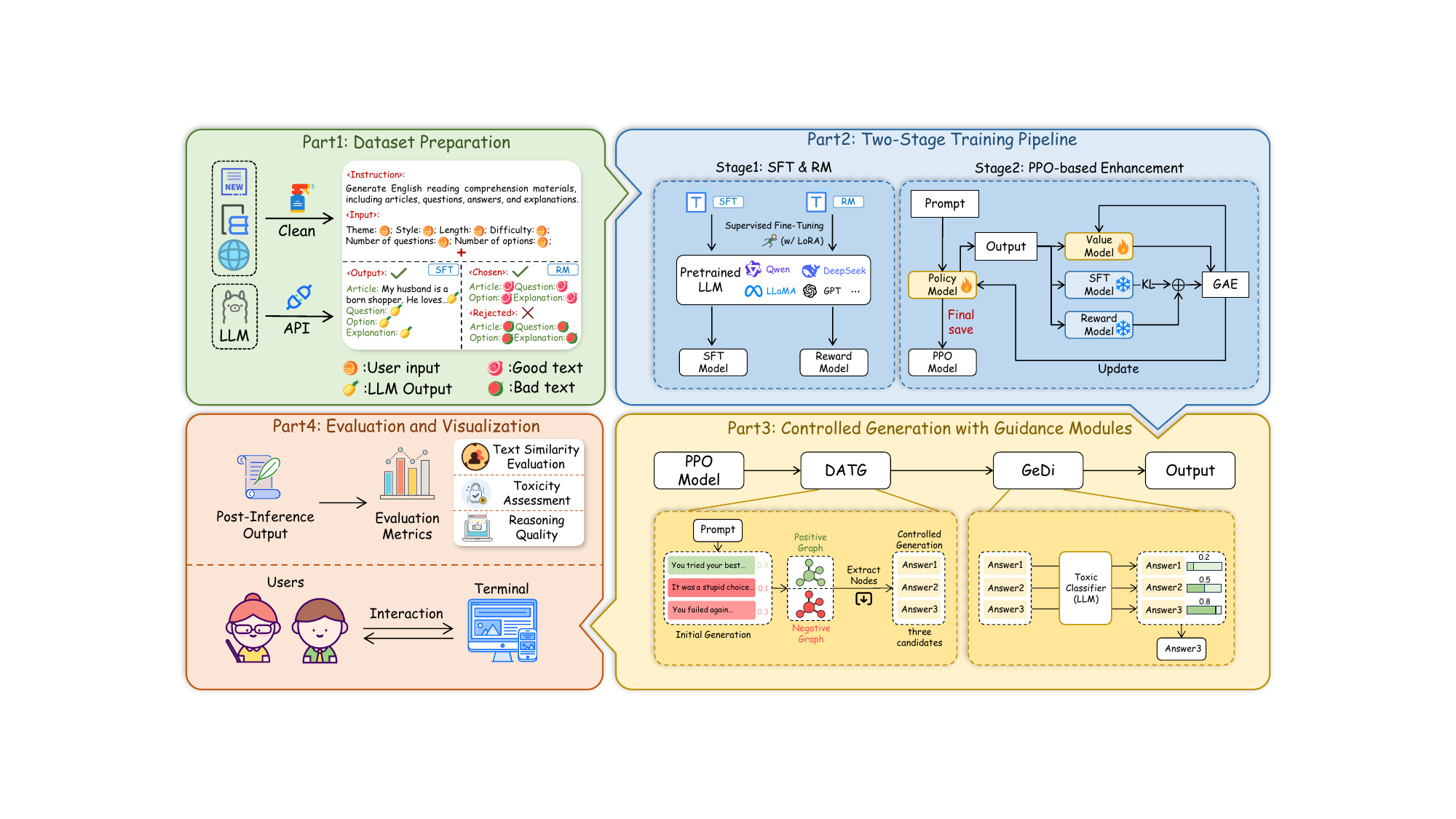}
\caption{Overall framework of Reading Comprehension Exercise Generation (RCEG), including dataset preparation, two-stage training pipeline, controlled generation with guidance modules, and evaluation and visualization.}
\label{fig2}
\end{figure}

Despite significant advances in large language models (LLMs), their use in education remains limited by poor task adaptability and limited control over output quality. To address this, we propose Reading Comprehension Exercise Generation (RCEG), a framework for automatically generating English reading comprehension materials. RCEG integrates instruction tuning, reward optimization, and post-generation control to improve pedagogical relevance and consistency. As illustrated in Fig.~\ref{fig2}, the framework consists of four core components: (i) dataset construction for supervised fine-tuning (SFT), reward modeling (RM), and PPO-based reinforcement learning; (ii) a two-stage training pipeline that improves instruction alignment and output quality; (iii) controlled generation using DATG and GeDi modules, enabling guided decoding and mitigation of linguistic toxicity; and (iv) evaluation and visualization modules that assess textual similarity, reasoning coherence, and user interaction outcomes.

\subsection{Two-stage Training Pipeline}
We design a two-stage training pipeline to improve the model’s adaptability to educational scenarios. This approach allows the model to better learn and align with the linguistic patterns and task structures commonly found in reading comprehension exercises.

\subsubsection{Supervised Fine-tuning and Reward Modeling}
In the first stage, we perform SFT using an instruction-style dataset specifically constructed for educational reading comprehension generation. Each training instance is formatted as a triplet $\tau = \left(\mathcal{I}, \mathcal{C}, \mathcal{R}\right)$, where $\mathcal{I}$ represents the instruction prompt (e.g., “Generate an English reading comprehension exercise, including the article, questions, answers, and explanations.”), $\mathcal{C}$ defines the content constraints, such as theme, style, word count, number of questions, number of options, and difficulty level, and $\mathcal{R}$ denotes the reference response, which consists of a complete reading comprehension exercise.

To enable parameter-efficient tuning, we adopt Low-Rank Adaptation (LoRA), which inserts trainable low-rank matrices into the attention layers while keeping the base model frozen. The training objective is to maximize the log-likelihood of the target response $\mathcal{R}$ conditioned on the instruction $\mathcal{I}$ and content constraints $\mathcal{C}$:

\begin{equation}
    \mathcal{L}_{\text{SFT}} = -\mathbb{E}_{\tau \sim \mathcal{D}} \sum_{t=1}^{T} \log \pi_\theta(r_t \mid \mathcal{I}, \mathcal{C}, r_{<t})
\end{equation}

\noindent where $\mathcal{D}$ denotes the training dataset, and $r_t$ represents the $t$-th token in the target response $\mathcal{R}$. This stage enables the model to learn how to follow instructional prompts and generate high-quality, pedagogically aligned content.

Meanwhile, we train a reward model to evaluate the quality of generated reading comprehension responses based on relevance and fluency. Each training instance consists of a shared input pair $\left(\mathcal{I}, \mathcal{C}\right)$ and two candidate outputs: a preferred response $\mathcal{R}^{+}$ and a less preferred response $\mathcal{R}^{-}$. These preference pairs are constructed by matching high-quality reference responses with lower-quality outputs generated by a baseline model using the DeepSeek-R1 API.

The reward model $r_\phi$ is trained to assign higher scores to preferred responses by optimizing a pairwise ranking objective:

\begin{equation}
    \mathcal{L}_{\text{RM}} = -\mathbb{E}_{(\mathcal{R}^{+}, \mathcal{R}^{-}) \sim \mathcal{D}_{\text{RM}}} \log \sigma \left( r_\phi(\mathcal{R}^{+}) - r_\phi(\mathcal{R}^{-}) \right)
\end{equation}

\noindent where $\sigma(\cdot)$ denotes the sigmoid function, and $\mathcal{D}_{\text{RM}}$ represents the reward model training dataset. This stage enables the reward model to function as an evaluator, providing feedback signals for subsequent reinforcement learning.

\subsubsection{Enhancement via Proximal Policy Optimization}
In the second stage, we apply PPO to further fine-tune the model based on feedback from the reward model. For each input pair $\left( \mathcal{I}, \mathcal{C} \right)$, the current policy $\pi_\theta$ generates a sampled response $\mathcal{R}$, which is then evaluated by the reward model. PPO aims to maximize a clipped surrogate objective:

\begin{equation}
    \mathcal{L}_{\text{PPO}} = \mathbb{E}_{\mathcal{R} \sim \pi_\theta} \left[ 
    \frac{1}{|\mathcal{R}|} \sum_{t=1}^{|\mathcal{R}|} \min \left( 
    r_t(\theta) \hat{A}_t,\ 
    \text{clip}\left(r_t(\theta), 1 - \epsilon, 1 + \epsilon\right) \hat{A}_t 
    \right) \right]
\end{equation}

\noindent where the importance weight $r_t(\theta)$ is defined as the ratio between the current policy $\pi_\theta$ and the old policy $\pi_{\theta_\text{old}}$:

\begin{equation}
    r_t(\theta) = \frac{\pi_\theta(r_t \mid \mathcal{I}, \mathcal{C}, r_{<t})}
    {\pi_{\theta_{\text{old}}}(r_t \mid \mathcal{I}, \mathcal{C}, r_{<t})}
\end{equation}

\noindent where $\hat{A}_t$ denotes the estimated advantage at time step $t$. To stabilize training and prevent over-optimization, we incorporate a KL-penalized reward at each token, computed as:

\begin{equation}
    \tilde{r}_t = r_\phi(\mathcal{I}, \mathcal{C}_{\le t}) - 
    \beta \log \left(
    \frac{\pi_\theta(r_t \mid \mathcal{I}, \mathcal{C}, r_{<t})}
    {\pi_{\text{ref}}(r_t \mid \mathcal{I}, \mathcal{C}, r_{<t})}
    \right)
\end{equation}

\noindent where $r_\phi$ denotes the reward model, and $\beta$ controls the strength of the KL penalty. This formulation encourages the model to produce higher-quality outputs while preserving alignment with the instruction-following behavior learned during the SFT stage.

\subsection{Controlled Generation with Guidance Modules}

\subsubsection{DATG-based Candidate Generation}
To enhance the controllability of generated outputs, we adopt a post-hoc mechanism based on DATG~\cite{liang2024controlled}. This module guides the generation process toward desirable textual attributes, such as non-toxicity, fluency, and pedagogical appropriateness.

Given an input prompt, the PPO model generates a baseline response. Around this baseline, we sample 30 candidate variants using stochastic decoding. Each variant $\tilde{R}_i$ is then evaluated by a pretrained attribute classifier $f_{\text{attr}}$, which assigns an alignment score $\mu_i = f_{\text{attr}}(\tilde{R}_i) \in [0, 1]$ to indicate its compatibility with the target attribute.

We then construct two directed token-level graphs: a positive graph $\mathcal{G}^+ = (\mathcal{V}, \mathcal{E}, w^+)$ and a negative graph $\mathcal{G}^- = (\mathcal{V}, \mathcal{E}, w^-)$, where the edge weights are defined as $w^+_{i,j} = \mu_i$ and $w^-_{i,j} = 1 - \mu_i$. These weights reflect the transition likelihoods associated with high or low alignment to the target attribute.

\noindent A ranking algorithm is then applied to identify the top 10 most influential tokens in each graph:

\begin{equation}
    \mathcal{V}_{\text{pos}} = \{ v \in \mathcal{V} \mid \text{Rank}_{\mathcal{G}^+}(v) > \delta_1 \}, \quad
    \mathcal{V}_{\text{neg}} = \{ v \in \mathcal{V} \mid \text{Rank}_{\mathcal{G}^-}(v) > \delta_2 \}
\end{equation}

\noindent To incorporate these token-level preferences into the generation process, we modify the model’s output logits at each decoding step. Specifically, the logits of positively ranked tokens are boosted, while those of negatively ranked tokens are suppressed:

\begin{equation}
    \tilde{P}(x_t \mid x_{<t}) = \text{softmax} \left( z_t + \lambda_1 \cdot \mathbf{m}_{\text{pos}} - \lambda_2 \cdot \mathbf{m}_{\text{neg}} \right)
\end{equation}

\noindent where $z_t$ denotes the original logits vector, and $\mathbf{m}_{\text{pos}}$ and $\mathbf{m}_{\text{neg}}$ are binary masks indicating the presence of selected positive and negative tokens, respectively. In our implementation, we set $\lambda_1 = 4.0$ and $\lambda_2 = 6.0$ to control the influence strength. The final decoding process leverages these adjusted probabilities to generate three attribute-guided candidates, denoted as $\tilde{R}_1$, $\tilde{R}_2$, and $\tilde{R}_3$, which are subsequently passed to the GeDi-based filtering module for final selection.

\subsubsection{GeDi-based Discriminative Filtering}
Following candidate generation by the DATG module, we perform a discriminative filtering stage using a pretrained toxic content classifier derived from the GeDi~\cite{krause2021gedi}. Instead of adopting the full GeDi generation pipeline, we leverage only the released toxic classifier to evaluate the final responses during post-processing.

The candidate responses $\tilde{R}_1$, $\tilde{R}_2$, and $\tilde{R}_3$ are evaluated by the classifier based on their toxicity levels. The final output, denoted as $\mathcal{R}_{\text{final}}$, is selected as the candidate with the highest non-toxicity score:

\begin{equation}
    R_{\text{final}} = \arg \max_{i \in \{1,2,3\}} \Bigl( s_i = f_{\textrm{GeDi}}(\tilde{R}_i) \Bigr), \quad s_i \in [0,1]
\end{equation}

\noindent This filtering strategy enables the incorporation of task-specific constraints without altering the generation process itself. It complements the DATG module by eliminating candidates that are stylistically or semantically misaligned, while preserving fluency and relevance.

\section{Experiments}
In this section, we conduct a series of experiments to evaluate the effectiveness of the proposed RCEG framework.

\subsection{Experimental Settings}
\subsubsection{Datasets}
We constructed three task-specific datasets for SFT, RM, and PPO to meet the requirements of each training stage. An overview of all datasets used in this study is presented in Table~\ref{tab1}.

\begin{table}[]
\centering
\caption{Construction details of the three datasets used for SFT, RM and PPO training in the RCEG framework.}
\label{tab1}
\renewcommand{\arraystretch}{1.3} 
\setlength{\tabcolsep}{6pt}
\begin{tabular}{c|c|c|c}
\hline
Dataset Type & Source Corpus & Sample Size & Augmentation Tool \\
\specialrule{1pt}{0pt}{0pt}
SFT          & RACE          & 24,636      & DeepSeek-R1       \\ \hline
RM           & Gaokao+PG     & 437         & DeepSeek-R1       \\ \hline
PPO          & OpenAI API    & 2,000       & OpenAI API        \\ \hline
\end{tabular}
\end{table}

\paragraph{Supervised Fine-tuning dataset}
We use the RACE~\cite{lai2017race} dataset as the foundation and select 24,636 high-quality samples based on content clarity and linguistic integrity. These samples are reformatted into the Alpaca-style instruction–output format to support LLMs fine-tuning. Since the original RACE dataset only includes the article, question, and answer, we extend each sample by adding an explanatory component. This addition is designed to help learners understand the rationale behind both correct and incorrect answer choices, thereby enhancing their comprehension skills. To generate these explanations, we utilize the DeepSeek-R1 API to produce fluent and pedagogically relevant rationales in Chinese for each sample. The resulting dataset forms the basis of the SFT stage in the RCEG framework.

\paragraph{Reward Modeling and Proximal Policy Optimization dataset}
To enable the reward model to distinguish between high and low quality outputs, we constructed a preference dataset for reinforcement learning. This dataset was built using reading comprehension passages collected from Gaokao English examinations (2015–present) and Postgraduate English Entrance Examinations (2010–present). Following the standard preference data format, each sample consists of a prompt paired with both a preferred and a rejected response. The rejected responses were generated using the DeepSeek-R1 API, producing lower-quality completions for each prompt. These responses, which lack coherence or pedagogical relevance, help the reward model learn to distinguish desirable qualities in generated text. This forms the foundation for subsequent PPO training.

In addition, we employed the OpenAI API to generate diverse completions for PPO data construction. These samples, paired with reward model scores, enable policy optimization via trial-and-error learning. Together, the preference and PPO datasets form the foundation for aligning model behavior with instructional objectives through reinforcement learning.

\subsubsection{Implementations}
We train our models using three curated datasets, with all fine-tuning performed using the LoRA method. The training configuration includes a batch size of 2, gradient accumulation steps of 8, and a learning rate of 1e-5. Both the SFT and PPO stages are trained for 3 epochs, while the RM is trained for 8 epochs. The cutoff length is truncated to 2048 tokens. All experiments are conducted on a single NVIDIA GeForce RTX 4090 GPU. All training procedures are implemented within the LLaMA-Factory framework~\cite{zheng2024llamafactory}.

\subsection{Evaluation Metrics}
We evaluate the generated outputs from multiple perspectives, including content quality, fluency, and safety. For content quality, we report BLEU~\cite{papineni2002bleu} and ROUGE~\cite{lin2004rouge} scores, which assess lexical overlap with reference texts based on n-gram precision and recall. Fluency is measured using perplexity (PPL)~\cite{liang2024controlled}, which reflects the model’s confidence in generating coherent and grammatically consistent text. Safety is assessed using a pretrained toxicity classifier applied during post-processing. This component is used to filter out harmful responses rather than for quantitative reporting, and therefore no separate safety scores are presented in the result tables. Model alignment under different prompt settings is evaluated using task-specific accuracy or classifier-based scores, as detailed in the corresponding result tables. The formulations of the three primary evaluation metrics are presented below:

\begin{equation}
    \text{BLEU} = \text{BP} \cdot \exp\left( \sum_{n=1}^{N} w_n \log p_n \right)
\end{equation}

\begin{equation}
    \text{ROUGE-N} = \frac{
    \sum_{\text{gram}_n \in \text{Ref}} \min\left(\text{Count}_{\text{gen}}, \text{Count}_{\text{ref}}\right)
    }{
    \sum_{\text{gram}_n \in \text{Ref}} \text{Count}_{\text{ref}}
    }
\end{equation}

\begin{equation}
    \text{PPL} = \exp\left( -\frac{1}{T} \sum_{t=1}^{T} \log P(x_t \mid x_{<t}) \right)
\end{equation}

\subsection{Results and Analysis}
We report model performance across four key dimensions: text similarity, reasoning alignment, fluency, and textual safety. BLEU and ROUGE results provide a surface-level comparison between model outputs and reference answers. Reasoning quality is evaluated using FreeEval~\cite{yu2024freeeval}, which quantifies alignment with instructional goals. Perplexity scores reflect language modeling fluency, while perplexity on toxic inputs serves as an indirect proxy for safety and robustness under adversarial conditions. We analyze model behavior under these dimensions and report detailed comparisons in the subsequent result tables.

\subsubsection{Text Similarity Evaluation}

\begin{table}[]
\caption{Performance comparison of baseline models and their RCEG-SP variants, which are optimized via SFT and PPO without post-processing. The evaluation is based on BLEU-4, ROUGE-1, ROUGE-2, and ROUGE-L metrics. Italic entries represent the RCEG-SP models.}
\label{tab2}
\centering
\renewcommand{\arraystretch}{1.3} 
\setlength{\tabcolsep}{6pt}       
\begin{tabularx}{\textwidth}{p{3.4cm}|>{\centering\arraybackslash}X|>{\centering\arraybackslash}X|>{\centering\arraybackslash}X|>{\centering\arraybackslash}X}
\hline
Models               & BLEU-4 & ROUGE-1 & ROUGE-2 & ROUGE-L \\ 
\specialrule{1pt}{0pt}{0pt}
GPT2\_finetune       & 1.39   & 12.42   & 1.10    & 4.80    \\ \hline
Qwen2.5-1.5B         & 12.86  & 16.60   & 4.15    & 11.34   \\ \hline
Qwen2.5-3B           & 24.24  & 21.80   & 5.93    & 14.82   \\ \hline
Qwen2.5-7B           & 38.01  & 25.51   & 4.72    & 12.45   \\ \hline
Qwen3-1.7B           & 25.84  & 23.89   & 7.08    & 13.88   \\ \hline
Qwen3-4B             & 31.22  & 24.95   & 6.78    & 14.28   \\ \hline
Llama-3.2-1B         & 1.45   & 6.08    & 0.82    & 3.71    \\ \hline
Llama-3.2-3B         & 1.36   & 7.13    & 0.28    & 2.91    \\ \hline
deepseek-llm-7b-base & 36.81  & 33.16   & 13.99   & 25.28   \\
\specialrule{1pt}{0pt}{0pt}
\textit{\textbf{RCEG-SP}}\textsubscript{\text{Qwen2.5-1.5B}} & \textbf{18.64} & \textbf{17.88} & \textbf{4.31} & 11.01          \\ \hline
\textit{\textbf{RCEG-SP}}\textsubscript{\text{Qwen2.5-3B}}   & \textbf{27.23} & \textbf{22.12} & \textbf{6.01} & \textbf{15.11} \\ \hline
\textit{\textbf{RCEG-SP}}\textsubscript{\text{Qwen2.5-7B}}   & \textbf{39.52} & 25.08          & \textbf{6.63} & \textbf{16.58} \\ \hline
\textit{\textbf{RCEG-SP}}\textsubscript{\text{Llama-3.2-3B}} & \textbf{27.53} & \textbf{23.83} & \textbf{5.03} & \textbf{11.35} \\
\hline
\end{tabularx}
\end{table}

Table~\ref{tab2} presents the text similarity evaluation results for the baseline models and their RCEG-SP variants, using BLEU and ROUGE metrics. All RCEG-SP models are optimized through SFT and PPO, without applying any post-processing modules.

Compared with the GPT2\_finetune model from prior work~\cite{xiao2023evaluating}, our RCEG-SP variants show substantial improvements across all evaluation metrics. For example, \textit{\textbf{RCEG-SP}}\textsubscript{Qwen2.5-3B} achieves a BLEU-4 score of 27.23 and a ROUGE-1 score of 22.12, significantly outperforming both its base model and the GPT2 baseline. These results show that combining SFT with PPO effectively improves lexical alignment between generated outputs and reference answers. This leads to higher-quality and more reliable text generation.

\subsubsection{Reasoning Quality Evaluation}
Table~\ref{tab3} presents the zero-shot evaluation results under the FreeEval benchmark, covering four representative reasoning tasks: ARC, ReClor, HotpotQA, and HellaSwag. All models are evaluated without task-specific tuning or in-context examples, reflecting their inherent reasoning capabilities.

The results show that RCEG-SP models achieve notable improvements on HotpotQA and HellaSwag, especially for smaller base models. For example, Qwen2.5-1.5B improves from 27.02 to 41.39 on HellaSwag, and Qwen2.5-3B reaches 93.34 on HotpotQA, exceeding its base variant by over 22 points. These results suggest better generalization to multi-hop and commonsense reasoning tasks in zero-shot settings. In contrast, performance on ARC and ReClor is mixed. While some models (e.g., Qwen2.5-3B, LLaMA-3B) remain stable, others (e.g., Qwen2.5-1.5B) show declines, indicating possible trade-offs in generalization across task types.

\begin{table}[]
\caption{Zero-shot FreeEval scores of base models and their RCEG-SP variants across four representative reasoning tasks: ARC, ReClor, HotpotQA, and HellaSwag. All models are evaluated without task-specific prompt tuning or in-context examples. Bold values indicate the better-performing variant for each model-task pair.}
\centering
\label{tab3}
\renewcommand{\arraystretch}{1.3} 
\setlength{\tabcolsep}{6pt}       
\begin{tabularx}{\textwidth}{p{2cm}|>{\centering\arraybackslash}X|>{\centering\arraybackslash}X|>{\centering\arraybackslash}X|>{\centering\arraybackslash}X|>{\centering\arraybackslash}X}
\hline
Models                        & Methods & ARC            & ReClor         & HotpotQA       & HellaSwag      \\ 
\specialrule{1pt}{0pt}{0pt}
\multirow{2}{*}{Qwen2.5-1.5B} & Base    & \textbf{64.24} & \textbf{47.53} & 89.25          & 27.02          \\ \cline{2-6} 
                              & RCEG-SP & 54.69          & 35.53          & \textbf{95.25} & \textbf{41.39} \\ \hline
\multirow{2}{*}{Qwen2.5-3B}   & Base    & 58.98          & 52.06          & 71.15          & 31.21          \\ \cline{2-6} 
                              & RCEG-SP & \textbf{63.53} & \textbf{52.06} & \textbf{93.34} & \textbf{33.62} \\ \hline
\multirow{2}{*}{Qwen2.5-7B}   & Base    & \textbf{84.61} & 65.8           & \textbf{95.35} & 29.67          \\ \cline{2-6} 
                              & RCEG-SP & 82.67          & \textbf{66.13} & 94.77          & \textbf{30.23} \\ \hline
\multirow{2}{*}{Llama-3.2-3B}     & Base    & 36.46          & 25.4           & \textbf{66.19} & 48.14          \\ \cline{2-6} 
                              & RCEG-SP & \textbf{36.47} & \textbf{26.4}  & 65.48          & \textbf{50.25} \\ \hline
\end{tabularx}
\end{table}

\subsubsection{Toxicity Assessment}

To evaluate model robustness under toxic or adversarial inputs, we report perplexity scores across three evaluation sets. All models are trained with SFT and PPO, followed by post-processing with DATG and GeDi to enable guided generation.

\begin{table}[]
\caption{Perplexity evaluation of models enhanced with post-processing on three datasets: SFT-TEST (test data at SFT), ToxicRandom (random toxic samples), and ToxicTop (highly toxic samples). All models are trained with SFT and PPO prior to post-processing. Bold values highlight improved performance over the base variant.}
\label{tab4}
\centering
\renewcommand{\arraystretch}{1.3} 
\setlength{\tabcolsep}{6pt}       
\begin{tabular}{c|c|c|c|c}
\hline
Models                                & Methods & SFT-TEST $\downarrow$  & ToxicRandom $\downarrow$  & ToxicTop $\downarrow$  \\ 
\specialrule{1pt}{0pt}{0pt}
\multirow{2}{*}{Qwen2.5-1.5B}         & Base    & 17.30          & 31.35          & 37.76          \\ \cline{2-5} 
                                      & RCEG & \textbf{9.16}  & \textbf{28.27} & \textbf{29.59} \\ \hline
\multirow{2}{*}{Qwen2.5-3B}           & Base    & 18.78          & 31.76          & 39.65          \\ \cline{2-5} 
                                      & RCEG & \textbf{13.39} & \textbf{30.23} & \textbf{30.17} \\ \hline
\multirow{2}{*}{Qwen2.5-7B}           & Base    & 20.74          & 33.24          & 40.95          \\ \cline{2-5} 
                                      & RCEG & \textbf{10.76} & \textbf{30.50} & \textbf{33.49} \\ \hline
\multirow{2}{*}{Llama-3.2-3B}         & Base    & 15.02          & 31.75          & 35.74          \\ \cline{2-5} 
                                      & RCEG & \textbf{8.10}  & \textbf{29.61} & \textbf{26.21} \\ \hline
\multirow{2}{*}{deepseek-llm-7b-base} & Base    & 18.51          & 37.85          & 42.79          \\ \cline{2-5} 
                                      & RCEG & \textbf{8.98}  & \textbf{34.34} & \textbf{30.86} \\ \hline
\end{tabular}
\end{table}

As shown in Table~\ref{tab4}, RCEG variants with post-processing consistently achieve lower perplexity across all subsets. For instance, Qwen2.5-1.5B drops from 17.30 to 9.16 on the SFT set and from 37.76 to 29.59 on the ToxicTop set. Similar gains are observed for Qwen2.5-3B and DeepSeek, indicating enhanced generation stability under varying input conditions.

These results suggest that the full RCEG pipeline not only improves fluency on clean data but also boosts model reliability when handling toxic inputs, contributing to overall textual safety.

\subsubsection{Interactive Interface}
\begin{figure}
\centering
\fbox{\includegraphics[width=0.98\textwidth]{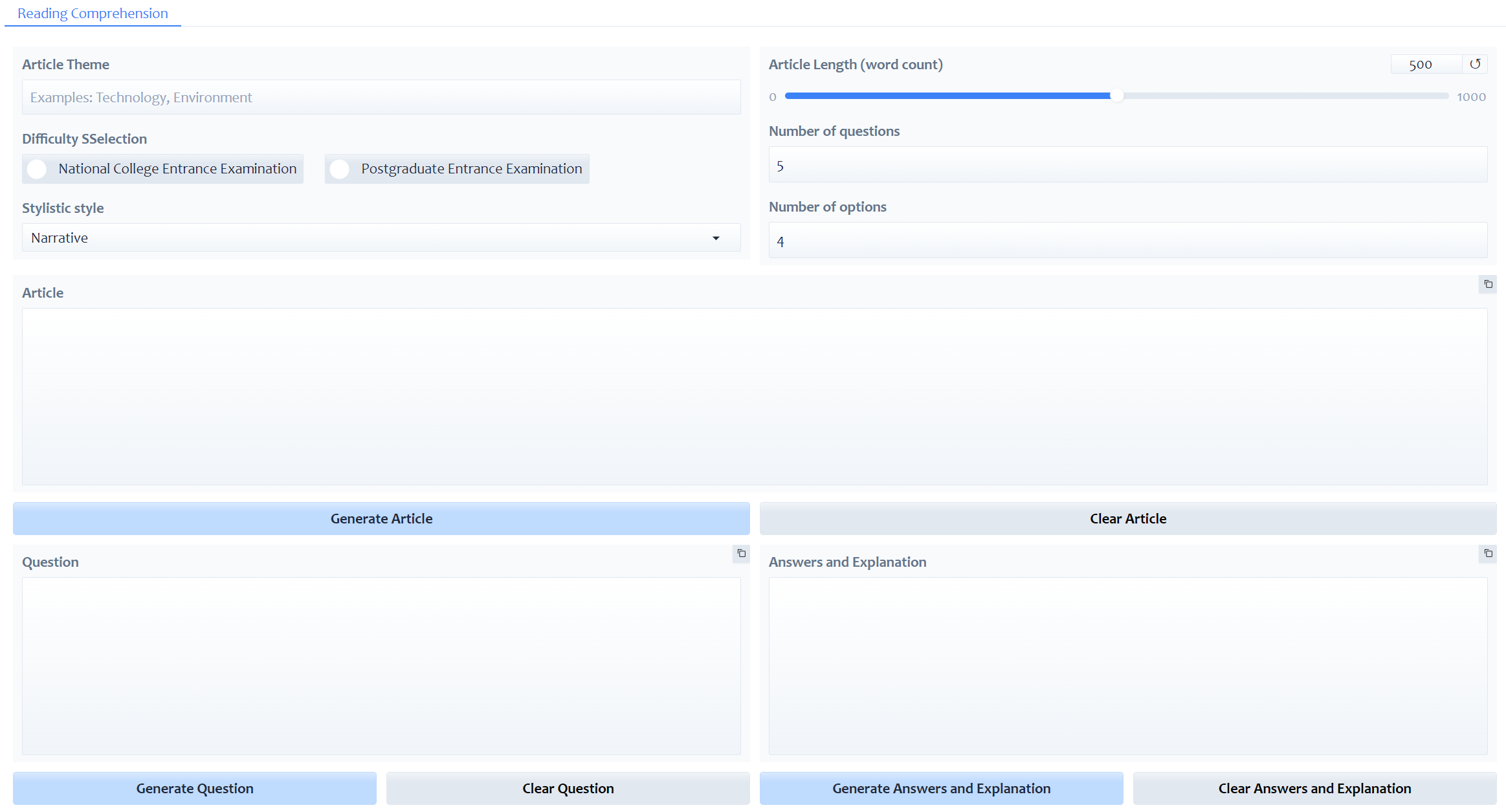}}
\caption{Gradio-based interactive interface for RCEG.}
\label{fig3}
\end{figure}

To facilitate practical use and intuitive inspection of model outputs, we developed a lightweight web-based interface using Gradio, as shown in Fig.~\ref{fig3}. The interface allows users to input topic and difficulty level, control the number of questions and options, and generate reading comprehension passages and corresponding questions and answers in real time. This demo system provides a convenient way to test the RCEG generation process and evaluate alignment quality from a human perspective.

\section{Conclusion}
This paper presents RCEG, a controllable generation framework tailored for educational reading comprehension content. RCEG leverages structured prompting and reward-aligned training to improve instruction adherence and content quality, without relying on external control modules. Extensive experiments across multiple language models show that RCEG-SP consistently improves BLEU, ROUGE, and task accuracy. When combined with post-hoc control, the complete RCEG pipeline further enhances fluency and robustness under diverse input conditions. These results demonstrate the scalability and effectiveness of RCEG in educational settings. In future work, we plan to extend the framework to multi-turn dialogue generation, domain adaptation, and real-world classroom deployment.

\subsubsection{Acknowledgement.}
This work is supported by the National Natural Science Foundation of China (No. 61603257).

\bibliographystyle{splncs04}
\bibliography{main}

@article{hao2024transforming,
  title={Transforming assessment: The impacts and implications of large language models and generative {AI}},
  author={Hao, Jiangang and von Davier, Alina A. and Yaneva, Victoria and Lottridge, Susan and von Davier, Matthias and Harris, Deborah J.},
  journal={Educational Measurement: Issues and Practice},
  volume={43},
  number={2},
  pages={pp. 16--29},
  year={2024},
  publisher={Wiley Online Library}
}

@article{voultsiou2025systematic,
  title={A systematic review of {AI}, {VR}, and {LLM} applications in special education: Opportunities, challenges, and future directions},
  author={Voultsiou, Evdokia and Moussiades, Lefteris},
  journal={Education and Information Technologies},
  pages={1--41},
  year={2025},
  publisher={Springer}
}

@article{kasneci2023chatgpt,
  title={{ChatGPT} for good? {On} opportunities and challenges of large language models for education},
  author={Enkelejda Kasneci and Kathrin Sessler and Stefan Küchemann and Maria Bannert and Daryna Dementieva and Frank Fischer and Urs Gasser and Georg Groh and Stephan Günnemann and Eyke Hüllermeier and others},
  journal={Learning and Individual Differences},
  volume={103},
  pages={pp. 102274},
  year={2023},
  publisher={Elsevier}
}

@article{wang2024large,
  title={Large language models for education: A survey and outlook},
  author={Wang, Shen and Xu, Tianlong and Li, Hang and Zhang, Chaoli and Liang, Joleen and Tang, Jiliang and Yu, Philip S and Wen, Qingsong},
  journal={arXiv preprint arXiv:2403.18105},
  year={2024}
}

@inproceedings{attali2018automatic,
  title={Automatic item generation unleashed: An evaluation of a large-scale deployment of item models},
  author={Attali, Yigal},
  booktitle={Artificial Intelligence in Education: 19th International Conference (AIED)},
  pages={17--29},
  year={2018},
}

@article{attali2022interactive,
  title={The interactive reading task: Transformer-based automatic item generation},
  author={Attali, Yigal and Runge, Andrew and LaFlair, Geoffrey T and Yancey, Kevin and Goodwin, Sarah and Park, Yena and Von Davier, Alina A},
  journal={Frontiers in Artificial Intelligence},
  volume={5},
  pages={pp. 903077},
  year={2022},
  publisher={Frontiers Media SA}
}

@article{bulut2022automatic,
  title={Automatic story and item generation for reading comprehension assessments with transformers},
  author={Bulut, Okan and Yildirim-Erbasli, Seyma Nur},
  journal={International Journal of Assessment Tools in Education},
  volume={9},
  number={Special Issue},
  pages={pp. 72--87},
  year={2022},
  publisher={{\.I}zzet KARA}
}

@inproceedings{liang2024controlled,
  title={Controlled Text Generation for Large Language Model with Dynamic Attribute Graphs},
  author={Liang, Xun and Wang, Hanyu and Song, Shichao and Hu, Mengting and Wang, Xunzhi and Li, Zhiyu and Xiong, Feiyu and Tang, Bo},
  booktitle={Findings of the Association for Computational Linguistics (ACL)},
  pages={5797--5814},
  year={2024}
}

@inproceedings{krause2021gedi,
  title={{GeDi}: Generative Discriminator Guided Sequence Generation},
  author={Krause, Ben and Gotmare, Akhilesh Deepak and McCann, Bryan and Keskar, Nitish Shirish and Joty, Shafiq and Socher, Richard and Rajani, Nazneen Fatema},
  booktitle={Findings of the Association for Computational Linguistics (ACL)},
  pages={4929--4952},
  year={2021}
}

@article{eddy1996hidden,
  title={Hidden {Markov} models},
  author={Eddy, Sean R},
  journal={Current Opinion in Structural Biology},
  volume={6},
  number={3},
  pages={pp. 361--365},
  year={1996},
  publisher={Elsevier}
}

@article{berger1996maximum,
  title={A maximum entropy approach to natural language processing},
  author={Berger, Adam and Della Pietra, Stephen A and Della Pietra, Vincent J},
  journal={Computational Linguistics},
  volume={22},
  number={1},
  pages={pp. 39--71},
  year={1996}
}

@article{sutton2012introduction,
  title={An introduction to conditional random fields},
  author={Sutton, Charles and McCallum, Andrew and others},
  journal={Foundations and Trends{\textregistered} in Machine Learning},
  volume={4},
  number={4},
  pages={pp. 267--373},
  year={2012},
  publisher={Now Publishers, Inc.}
}

@article{iqbal2022survey,
  title={The survey: Text generation models in deep learning},
  author={Iqbal, Touseef and Qureshi, Shaima},
  journal={Journal of King Saud University-Computer and Information Sciences},
  volume={34},
  number={6},
  pages={pp. 2515--2528},
  year={2022},
  publisher={Elsevier}
}

@article{li2024pre,
  title={Pre-trained language models for text generation: A survey},
  author={Li, Junyi and Tang, Tianyi and Zhao, Wayne Xin and Nie, Jian-Yun and Wen, Ji-Rong},
  journal={ACM Computing Surveys},
  volume={56},
  number={9},
  pages={pp. 1--39},
  year={2024},
  publisher={ACM New York, NY}
}

@inproceedings{sutskever2014sequence,
  title={Sequence to sequence learning with neural networks},
  author={Sutskever, Ilya and Vinyals, Oriol and Le, Quoc V},
  booktitle={Advances in Neural Information Processing Systems (NIPS)},
  pages={3104--3112},
  year={2014}
}

@inproceedings{vaswani2017attention,
  title={Attention is all you need},
  author={Vaswani, Ashish and Shazeer, Noam and Parmar, Niki and Uszkoreit, Jakob and Jones, Llion and Gomez, Aidan N and Kaiser, \L ukasz and Polosukhin, Illia},
  booktitle={Advances in Neural Information Processing Systems (NIPS)},
  pages={5998--6008},
  year={2017}
}

@article{zhao2023survey,
  title={A survey of large language models},
  author={Zhao, Wayne Xin and Zhou, Kun and Li, Junyi and Tang, Tianyi and Wang, Xiaolei and Hou, Yupeng and Min, Yingqian and Zhang, Beichen and Zhang, Junjie and Dong, Zican and others},
  journal={arXiv preprint arXiv:2303.18223},
  year={2023}
}

@inproceedings{chan2022agree,
  title={{AGReE}: A system for generating Automated Grammar Reading Exercises},
  author={Chan, Sophia and Somasundaran, Swapna and Ghosh, Debanjan and Zhao, Mengxuan},
  booktitle={Proceedings of the 2022 Conference on Empirical Methods in Natural Language Processing (EMNLP)},
  pages={169--177},
  year={2022}
}

@article{zu2023automated,
  title={Automated distractor generation for fill-in-the-blank items using a prompt-based learning approach},
  author={Zu, Jiyun and Choi, Ikkyu and Hao, Jiangang},
  journal={Psychological Testing and Assessment Modeling},
  volume={65},
  number={2},
  pages={pp. 55--75},
  year={2023}
}

@inproceedings{xiao2023evaluating,
  title={Evaluating reading comprehension exercises generated by {LLMs}: A showcase of {ChatGPT} in education applications},
  author={Xiao, Changrong and Xu, Sean Xin and Zhang, Kunpeng and Wang, Yufang and Xia, Lei},
  booktitle={Proceedings of the 18th workshop on innovative use of NLP for Building Educational Applications (BEA)},
  pages={610--625},
  year={2023}
}

@inproceedings{lai2017race,
  title={{RACE}: Large-scale ReAding Comprehension Dataset From Examinations},
  author={Lai, Guokun and Xie, Qizhe and Liu, Hanxiao and Yang, Yiming and Hovy, Eduard},
  booktitle={Proceedings of the 2017 Conference on Empirical Methods in Natural Language Processing (EMNLP)},
  pages={785--794},
  year={2017}
}

@inproceedings{zheng2024llamafactory,
  title={{LlamaFactory}: Unified Efficient Fine-Tuning of 100+ Language Models},
  author={Zheng, Yaowei and Zhang, Richong and Zhang, Junhao and YeYanhan, YeYanhan and Luo, Zheyan},
  booktitle={Proceedings of the 62nd Annual Meeting of the Association for Computational Linguistics (ACL)},
  pages={400--410},
  year={2024}
}

@inproceedings{papineni2002bleu,
  title={{BLEU}: A method for automatic evaluation of machine translation},
  author={Papineni, Kishore and Roukos, Salim and Ward, Todd and Zhu, Wei-Jing},
  booktitle={Proceedings of the 40th annual meeting of the Association for Computational Linguistics (ACL)},
  pages={311--318},
  year={2002}
}

@inproceedings{lin2004rouge,
  title={{ROUGE}: A package for automatic evaluation of summaries},
  author={Lin, Chin-Yew},
  booktitle={Text Summarization Branches Out},
  pages={74--81},
  year={2004}
}

@inproceedings{yu2024freeeval,
  title={{FreeEval}: A Modular Framework for Trustworthy and Efficient Evaluation of Large Language Models},
  author={Yu, Zhuohao and Gao, Chang and Yao, Wenjin and Wang, Yidong and Zeng, Zhengran and Ye, Wei and Wang, Jindong and Zhang, Yue and Zhang, Shikun},
  booktitle={Proceedings of the 2024 Conference on Empirical Methods in Natural Language Processing (EMNLP)},
  pages={1--13},
  year={2024}
}

\end{document}